\documentclass[runningheads]{llncs}
\usepackage{graphicx}
\usepackage{xcolor}
\usepackage{siunitx}
\usepackage{bm}
\usepackage{amsmath}
\usepackage{amssymb}
\usepackage{booktabs}
\usepackage[linesnumbered,ruled]{algorithm2e}
\usepackage{mathtools}
\usepackage{url}

\DeclareMathOperator*{\argmax}{arg\,max}

\newcommand{\DSC}{\mathcal{D}}
\newcommand{\NCC}{\mathcal{C}}
\newcommand\numberthis{\addtocounter{equation}{1}\tag{\theequation}}

\begin{document}
\title{3D Reconstruction and Segmentation of Dissection Photographs for MRI-free Neuropathology\\
\textit{(accepted at MICCAI 2020)}}
\titlerunning{3D Reconstruction and Segmentation of Dissection Photographs}

\author{Henry~F.~J.~Tregidgo\inst{1*} \and
Adri\`a~Casamitjana\inst{1} \and
Caitlin~S.~Latimer\inst{2} \and
Mitchell~D.~Kilgore\inst{2} \and
Eleanor~Robinson\inst{1} \and
Emily~Blackburn\inst{3} \and
Koen~Van~Leemput\inst{4,5} \and
Bruce~Fischl\inst{4} \and
Adrian~V.~Dalca\inst{4,6} \and
Christine~L.~Mac~Donald\inst{7} \and
C.~Dirk~Keene\inst{2} \and
Juan~Eugenio~Iglesias\inst{1,4,6} } 
%index{Tregidgo, Henry~F.~J.}
%index{Casamitjana, Adri\`a}
%index{Latimer, Caitlin~S.}
%index{Kilgore, Mitchell~D.}
%index{Robinson, Eleanor}
%index{Blackburn, Emily}
%index{Van~Leemput, Koen}
%index{Fischl, Bruce}
%index{Dalca, Adrian~V.}
%index{Mac~Donald, Christine~L.}
%index{Keene, C.~Dirk}
%index{Iglesias, Juan~Eugenio}

\authorrunning{H. Tregidgo et al.} % commented out for double blind review
\institute{Centre for Medical Image Computing, University College London, London, UK\\
\email{h.tregidgo@ucl.ac.uk} \and
Department of Pathology, University of Washington, Seattle, WA, USA\and
Queen Square Institute of Neurology, University College London, London, UK \and
Martinos Center for Biomed. Imaging, MGH \& Harvard Med. School, Boston, USA \and
Department of Health Technology, DTU, Lyngby, Denmark \and
Computer Science and Artificial Intelligence Laboratory, MIT, Cambridge, USA\and
Department of Neurological Surgery, University of Washington, Seattle, WA, USA}
% \institute{Institute one address \and
% Institute two address \and
% Institute three address \and
% Institute four address \and
% Institute five address\\
% \email{first.author@email.com}}
%
\maketitle           
%
%\vspace{-4pt}
\begin{abstract}
Neuroimaging to neuropathology correlation (NTNC) promis-es to enable the transfer of microscopic signatures of pathology to \emph{in vivo} imaging with MRI, ultimately enhancing  clinical care.  
NTNC traditionally requires a volumetric MRI scan, acquired either \emph{ex vivo} or a short time prior to death.
Unfortunately, \emph{ex vivo} MRI is difficult and costly, and recent \emph{premortem} scans of sufficient quality are seldom available. To bridge this gap, we present methodology to 3D reconstruct and segment full brain image volumes from  brain dissection photographs, which  are routinely acquired at many brain banks and neuropathology departments.  The 3D reconstruction is achieved via a joint registration framework, which uses a reference volume other than MRI. This volume may  represent either the sample at hand (e.g., a surface 3D scan) or the general population (a probabilistic atlas).  In addition, we present a  Bayesian method to segment the 3D reconstructed photographic volumes into 36 neuroanatomical structures, which is robust to nonuniform brightness within and across photographs. We evaluate our methods on a dataset with 24 brains, using Dice scores and volume correlations. The results show  that dissection photography is a valid replacement for \emph{ex vivo} MRI in many volumetric analyses,  opening an avenue for MRI-free NTNC, including retrospective data. The code is  available at \url{https://github.com/htregidgo/DissectionPhotoVolumes}.

% \keywords{First keyword  \and Second keyword \and Another keyword.}
\end{abstract}

\section{Introduction}

A crucial barrier to the study of  neurodegenerative diseases (e.g., Alzheimer's disease and its mimics~\cite{Nelson:2019aa,Armstrong:2005aa}) is the lack of reliable \emph{premortem} biomarkers, as definitive diagnoses can only be obtained via neuropathology. To overcome this, neuroimaging to neuropathology correlational (NTNC) science seeks to establish imaging phenotypes that correlate with gold standard pathological diagnoses, in order to port these signatures to \emph{in vivo} imaging as biomarkers. 
%Despite its great potential, NTNC remains largely unexplored.
One candidate for NTNC is \emph{in vivo} or \emph{ex vivo} MRI. Reliable matching of histology and \emph{in vivo} MRI requires a \emph{premortem} scan acquired a short time before death. Unfortunately, these are difficult to obtain for precisely the most interesting individuals -- asymptomatic, early-stage cases. This problem can be overcome with \emph{ex vivo} MRI, which has been successfully used in NTNC (e.g.,~\cite{Dawe:2011aa,Kotrotsou:2015aa}), but is also challenging to perform:  it requires scanning and sample preparation expertise that is not present at many research centres, cannot easily be done on the frozen tissue required in many genetics analyses, and is expensive.

Meanwhile, a wealth of information exists in brain banks that is hidden in existing images from routine dissection photography. 
Here we present algorithms to 3D reconstruct and segment imaging volumes from this underutilised modality, enabling morphometric NTNC studies without MRI at almost no cost. 
% %Based on our  experiments, we anticipate that the proposed tools will provide access to a wealth of information currently present but unusable in existing brain bank images from routine dissection photography.

%\vspace{4pt}
\noindent\textbf{Related work:} Building 3D images from dissection photographs requires alignment of a stack of 2D photographs into a 3D consistent volume via image registration~\cite{Maintz:1998aa,Sotiras:2013aa}.
Registration of image pairs is a well-studied problem but, to the best of our knowledge, literature on joint registration of dissection photographs for 3D reconstruction is nonexistent. 
The closest related work is a method for volumetric reconstruction from printed films of MRI~\cite{Ebner:2018aa}, which is not suitable for our task, as it requires a reference MRI volume (which we wish to avoid). 

One step further removed is 3D histology reconstruction~\cite{Pichat:2018aa}.
Despite the peculiarities of histological data in terms of contrast, resolution, and sectioning distortions, many of the challenges we face in this work are similar. Without an additional reference, recovering the 3D shape of a 2D stack of images is a heavily underconstrained problem. 
A common approach is to iteratively align each 2D image to its neighbours, possibly with an outlier rejection strategy~\cite{Yushkevich:2006aa}. 
This approach yields smooth volumes but can result in straightening of curved structures (often known as ``banana effect''~\cite{Yang:2012aa}), and accumulation of errors along the stack (\emph{z-shift},~\cite{Pichat:2017aa}). 
These can both be overcome with a reference MRI scan~\cite{amunts2013bigbrain,adler2014histology} -- a requirement which, again, we are trying to avoid.  

For brain segmentation, the neuroimaging literature has long been dominated by multi-atlas segmentation (MAS), Bayesian segmentation and, more recently, deep convolutional neural networks (CNNs). 
MAS~\cite{rohlfing2004evaluation,Iglesias:2015aa} 
nonlinearly registers several labelled atlases to a target scan, deforms the corresponding segmentations, and merges these warped label maps into a robust estimate of the segmentation with a label fusion algorithm. 
Segmentation CNN architectures~\cite{milletari_v-net_2016,kamnitsas_efficient_2017,litjens2017survey}, best represented by the ubiquitous U-Net~\cite{Ronneberger:2015aa}, yield state-of-the-art accuracy and runtimes (seconds). 
Being supervised methods, MAS and CNNs share the disadvantage that performance quickly decreases when the training and test domains do not match.
Despite progress in data augmentation~\cite{ZhaoDalcaCVPR2019} and transfer learning~\cite{shin2016deep}, manual labels are often needed for every new segmentation task.

Bayesian segmentation with probabilistic atlases uses a generative model combining  a supervised prior model of anatomy (the atlas) and a model of image formation (likelihood). Segmentation is then posed as  a Bayesian inference problem, estimating the most probable hidden segmentation that generated the observed image, given the atlas. A subset of Bayesian segmentation methods~\cite{van_leemput_automated_1999,zhang_segmentation_2001,Ashburner:2005aa,Puonti:2016aa} use an unsupervised likelihood model, usually a Gaussian mixture model (GMM) whose parameters are estimated specifically for each volume to segment, making the segmentation adaptive to different contrasts. 

%\vspace{4pt}
\noindent\textbf{Contribution:} 
The contribution of this paper is twofold.
First, we propose a joint registration algorithm for reconstruction of 3D imaging volumes from stacks of dissection photographs, requiring only a reference mask.  
This mask can be ``hard", e.g., measured directly with a 3D surface scanner~\cite{Geng:2011aa}, or ``soft", e.g., a probabilistic atlas of the whole brain; neither of these require MRI acquisition. 
Second, we present a Bayesian  algorithm to segment 3D reconstructed stacks of photographs into 36 brain structures. We have designed an unsupervised likelihood term that models photography-specific artifacts, readily adapting to photographic hardware and brain fixation differences,  thus making our publicly available code immediately usable with data from any institution.

\section{Methods}
\label{sec:methods}

\subsection{General workflow}
\label{sec:preprocessing}

\begin{figure}[t]
\centering
\includegraphics[width=0.90\textwidth]{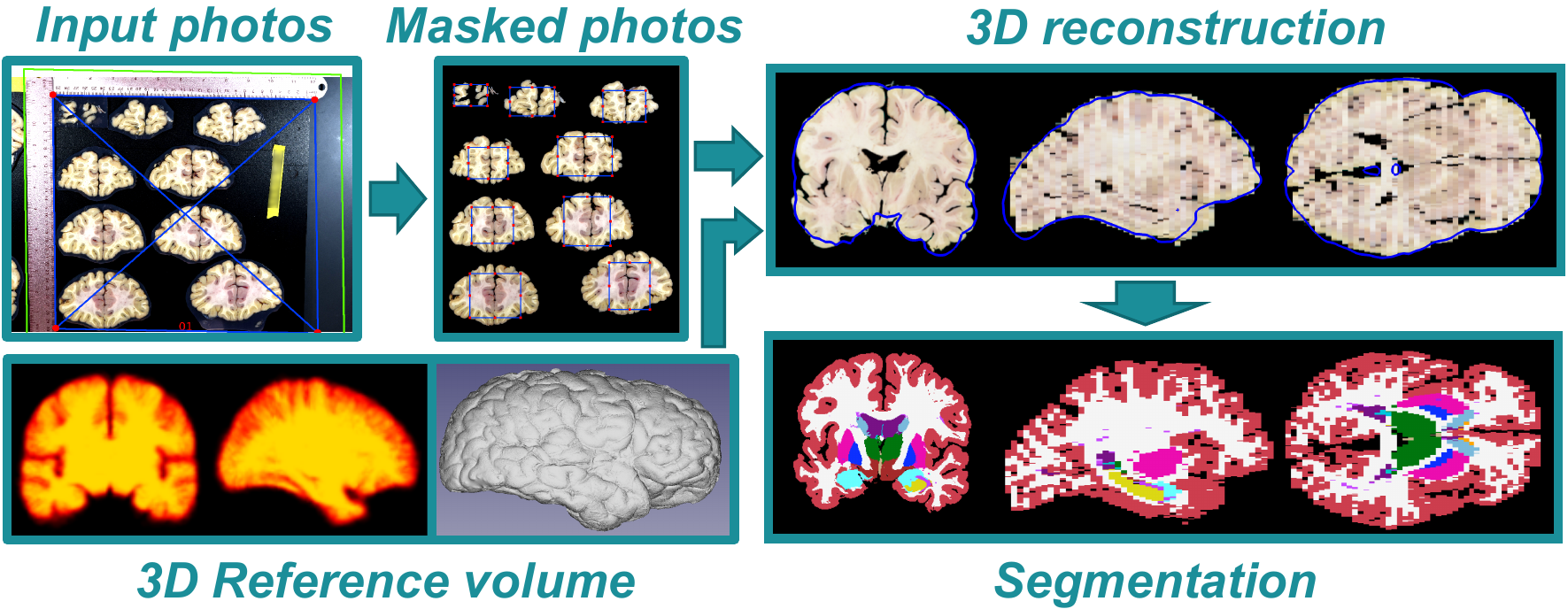}
%\vspace{-5pt}
\caption{
Diagram of the proposed processing pathway for dissection photography. 
Photographs are coarsely corrected for pixel size and perspective; arranged into a volume; registered to each other and a reference surface mask; and automatically segmented.}\label{fig:workflow} 
%\vspace{-5pt}
\end{figure}

The workflow of our framework is outlined in Fig.~\ref{fig:workflow}.
The inputs of the algorithm are dissection photographs of brain slices, and a reference volume~$R$ describing the exterior shape of the brain. In addition, the user provides two inputs. First, three landmarks on the photographs (e.g., on rulers which are commonly used in dissection photographs), which we use for coarse pixel size calibration and perspective correction. Second, segmentations for the different slices, which isolate tissue from background and encode the order of the slices. These can often be obtained with minimal interaction -- we use a simple GMM~\cite{Zivkovic:2004aa} requiring one click per slice.
The resulting slices and masks are ordered manually and arranged into sets of stacked slices $S= \{S_n\},$ and corresponding masks $M=\{M_n\}.$ %We will refer to the volumetric reference (hard or soft) as $R$. 

\subsection{3D reconstruction from dissection photographs}
\label{sec:photoReg}

Let $\{\Phi_n\}$ be a set of 2D affine geometric transforms for the brain slices, which brings them into alignment. These transforms correct for slice displacement and rotation, as well as perspective distortion. Then,  $S_{n}[\bm{x};\Phi_n]$ and $M_{n}[\bm{x};\Phi_n]$ denote a resampling of slice $n$ and its corresponding mask, to a discretised grid given by $\bm{x},$ where the in-plane coordinates have been transformed according to the parameters in $\Phi_n$. Similarly, we define a 3D rigid transform for the reference volume $R$ that brings it into alignment with the stack of slices, such that $R[\bm{x};\Psi]$ is a resampling of $R$ parameterised by the 3D transform in $\Psi$. If the reference $R$ is hard and directly represents the target shape, we also include an additional scaling in the direction of slicing 
(typically the anterior-posterior axis, for coronal slices) 
to account for deviations from the nominal slice thickness; with a soft atlas as reference, this is not possible.
We jointly register the reference volume and the slices by maximising the following objective function~$\mathcal{F}$:
\begin{align*}
\mathcal{F}(\Psi,\{\Phi_n\}) =  \;&\alpha\DSC(M[\bm{x};\{\Phi_n\}],R[\bm{x};\Psi]) + \beta \frac{1}{N_s} \sum_{n=1}^{N_s-1}\NCC(S_{n}[\bm{x};\Phi_n],S_{n+1}[\bm{x};\Phi_{n+1}]) \\
+& \gamma \frac{1}{N_s} \sum_{n=1}^{N_s-1}\DSC(M_n[\bm{x};\Phi_n],M_{n+1}[\bm{x};\Phi_{n+1}])  - \nu \frac{1}{N_s} \sum_{n=1}^{N_s}f(\Phi_n), \numberthis\label{eqn:cost}
\end{align*}
where $N_s$ is the number of slices, $\DSC$ is the Dice score, $\NCC$ is the normalised cross correlation, $f$ is a regulariser, and $\{\alpha,\beta,\gamma,\nu\}$ are relative weights for each term.

Equation~\ref{eqn:cost} corrects for overall shape by encouraging a high Dice similarity coefficient between $M[\bm{x};\{\Phi_n\}]$ and $R$, i.e., the 3D reconstructed mask and the reference volume.  Smoothness within the reconstructed photography volume $S[\bm{x};\{\Phi_n\}]$ is encouraged with two terms: the normalised cross correlation  between successive slices in $S$, and the Dice coefficient between the corresponding masks. The final term, $f(\Phi_n),$ is a regulariser used to constrain the 2D spatial transforms not to be excessively scaled or sheared, which is particularly useful for the first and last slices in the stack, as they often contain little tissue.

The registration is solved in a hierarchical fashion using two levels with increasing complexity, combined with a multi-scale approach, to help avoid local maxima and increase convergence speed.
At the first level, we limit the registration to correcting for slice displacement and rotation only, which is achieved  by constraining each $\Phi_n$ to be rigid rather than affine. At this level, no regularisation is needed. At the second level, we use the full model (i.e., affine $\{\Phi_n\}$). Undue scaling or shearing is avoided at this level by penalising transformations that excessively modify the area of a pixel, with $f(\Phi_n) = \big| \log | \Phi_n | \big|$ in Equation~\ref{eqn:cost}. 

Optimisation is performed at three levels of resolution (1/4, 1/2 and 1) with the L-BFGS algorithm~\cite{Byrd:1995aa,Nocedal:2006aa}. The 2D transforms are initialised by aligning the centres of gravity (COGs) of the masks, and the 3D transform  by matching the COG of the 3D mask with the COG of the initialised stack. 
Model parameters were set via visual inspection on a separate dataset. 
If the reference $R$ is soft, we set $\alpha=10$, $\beta=1$, $\gamma=2$, $\nu=0.1$. If $R$ is hard (i.e., measured directly), we give it a higher weight in the reconstruction ($\alpha=50$) and regularise less ($\nu=0.05$). 
%These parameters were set via visual inspection on a separate dataset. 

%An overview of the method is shown in Algorithm~\ref{alg:registration}.

%\begin{algorithm}[t]
%\caption{Photo volume registration}\label{alg:registration}
%
%    \SetKwInOut{Input}{Input}
%    \SetKwInOut{Output}{Output}
%    
%    \Input{Set of photo slices $S=\{S_n\}$, corresponding slice masks $M=\{M_n\}$ and reference volume $R$.}
%    \Output{Registered photo volume $V$ and registered reference volume $R$}
%    
%    Build resolution pyramid by defining $N_r$ discretisation grids $\bm{x_i}$ for resampling. % resampling slices, masks and reference volume to $N_r$ resolutions $S^i,M^i,B^i$~for~$i=1,\dots,N_r$%\vspace{2pt}
%    
%    Set initial level 1 registration parameters, $\Theta^1_0 = [\Psi^1_0,\{\Phi_n\}^1_0]$ by aligning COGs. %\vspace{2pt}
%
%    \For{$i=1:N_r$}
%    {
%        Update $\Theta^1_{i-1}$ to $\Theta^1_{i}$ using Eqn.~\ref{eqn:mode1} optimised with L-BFGS
%    }%\vspace{2pt}
%    
%    Set initial level 2 registration parameters, $\Theta^2_0 = [\Psi^2_0,\{\Phi_n\}^2_0]$ from $\Theta^1_{N_r}$%\vspace{2pt}
%        
%    \For{$i=1:N_r$}
%    {
%        Update $\Theta^2_{i-1}$ to $\Theta^2_{i}$ using Eqn.~\ref{eqn:mode2} optimised with L-BFGS
%    }%\vspace{2pt}
    
%    Set $V=S[\bm{x_{N_r}};\{\Phi_n\}^2_{N_r}]$ and $B=R[\bm{x_{N_r}};\Psi^2_{N_r}]$
    
%\end{algorithm}

\subsection{Segmentation}
\label{sec:segmentation}
The ultimate purpose of the photographic volumes is for morphometric analyses, most of which require image segmentation. Since our goal is to make our code available to other researchers, supervised CNNs or MAS may not be appropriate as they may not generalise well to photographs of brains that have been fixed with potentially very different protocols. Instead, we propose a Bayesian algorithm with an unsupervised likelihood that includes a model of artefacts specific to photography,  and thus adapts to cases fixed and imaged with any protocol. 

Specifically, we maximise the probability of a 3D label map $L$ given the image data $D$ using  Bayes' rule $p(L | D )\propto p(D | L)p(L)$. Both the prior and the likelihood have an associated set of parameters, $\theta_L$ and $\theta_D$, respectively, with prior distributions $p(\theta_L)$ and $p(\theta_D)$. The prior $p(L|\theta_L)$ is a publicly available probabilistic atlas of anatomy~\cite{Puonti:2016aa} with $K=36$ neuroanatomical classes, encoded as a tetrahedral mesh endowed with a deformation model~\cite{Van-Leemput:2009aa}. Each voxel of the segmentation is assumed to be an independent sample of the discrete distribution defined by the deformed atlas at the  location of the voxel. 

The likelihood $p(D|L,\theta_D)$ combines a GMM  with a model for brightness variations. Specifically, each  of the $K$ classes has an associated set of GMM parameters (weights, means, covariances), such that the intensity of a voxel is assumed to be a sample of the GMM associated with its label. These intensities are further corrupted by a slice-specific, smooth, multiplicative field (henceforth ``brightness field''), which we assume to be a linear combination of smooth basis functions allowing bilinear variation in plane, independently for each slice.

It is typical in Bayesian segmentation to first compute point estimates of the model parameters $(\hat\theta_L,\hat\theta_D)$ by maximising $p(\theta_L,\theta_D|D)$, and then to estimate the segmentation as the maximum of $p(D | L,\hat\theta_D) p(L|\hat\theta_L)$. 
Let $\bm{\Gamma}=\{\bm{\Gamma}_k\}_{k=1}^K$ be the GMM parameters of the different classes, $\bm{C}$  the matrix of brightness field coefficients (with 3 rows and as many columns as basis functions), and $\bm{x}$ the atlas mesh position, such that $\theta_D=(\bm{\Gamma},\bm{C})$ and $\theta_L=\bm{x}$. Then, taking the logarithm of $p(\theta_L,\theta_D | D)$ yields the following objective function for the model parameters:
\begin{small}
\begin{equation}
\left\{\bm{\hat{x}},\hat{\bm{\Gamma}},\hat{\bm{C}}\right\} = \argmax_{\bm{x},\bm{\Gamma},\bm{C}} \sum_{i=1}^N\log\left(\sum_{k=1}^K p_i(\bm{d}_i|\bm{\Gamma},\bm{C},k) p_i(k|\bm{x})\right)+\log p(\bm{x}) + \log p(\bm{C}), 
\label{eqn:samsegcost}
\end{equation}
\end{small}
where  $d_i$ is the vector with the  log-transformed RGB intensities of voxel $i$  and $N$ is the number of voxels; a flat prior is assumed for  $\bm{\Gamma}$.
The likelihood term is: 
$$
p_i(\bm{d}_i|k,\bm{C},\bm{\Gamma}) = \sum_{g=1}^{G_k}w_{k,g}\mathcal{N}(\bm{d}_i-\bm{C}\bm{\phi}_i|\bm{\mu}_{k,g},\bm{\Sigma}_{k,g}),
$$
where $G_k$ is the number of components of the GMM of class $k$;  $w_{k,g}$ is the weight of component $g$ of class $k$; $\mathcal{N}$ is the Gaussian distribution; $\bm{\phi}_i$ is a vector with the values of the brightness field basis functions at voxel $i$; and $\bm{\mu}_{k,g},\bm{\Sigma}_{k,g}$ are the mean vector and covariance matrix associated with component $g$ of class $k$.

Segmentation is achieved by maximising Equation~\ref{eqn:samsegcost} with coordinate ascent. We numerically optimise $\theta_L$  with L-BFGS, initialised with an affine transform computed by registering the atlas to $R$ with a robust approach~\cite{Reuter:2010aa}. We optimise $\theta_D$ with the Generalised EM algorithm~\cite{Dempster:1977aa}. GEM involves iteratively: \textit{(i)}~constructing a lower bound of the objective function that touches it at the current estimate of the parameters, which amounts to a soft classification of each voxel (E step); and \textit{(ii)}~improving this bound to update $\theta_D$ (generalised M step). Upon convergence, the probabilistic segmentation $p(D | L,\hat\theta_D) p(L|\hat\theta_L)$ is given by the soft classification of the final E step. 
We implement this optimisations by adapting routines from the public SAMSEG repository~\cite{Puonti:2016aa}. 
All parameters (number of Gaussians $G_k$, mesh stiffness, etc.) are set to default SAMSEG values.

\section{Experiments and Results}
\label{sec:results}

\subsection{Datasets}
\label{sec:data}

We used a dataset consisting of dissection photography and matched \emph{ex vivo} MRI for 24 cases, including only the cerebrum (i.e., no cerebellum or brainstem).
Photographs were acquired of slices cut in the coronal plane with \SI{4}{mm} thickness,  using a 35 Megapixel camera.
After initial perspective and pixel size adjustments, the calibrated photos were resampled to a pixel size of \SI{0.1}{mm}.
The \emph{ex vivo} MRI were acquired using a FLAIR sequence with \SI{0.8}{mm} isotropic resolution. 

To evaluate the algorithms, we used two sets of reference segmentations. The first set consists of sparse manual delineations made on one slice photograph per volume. The slices were chosen to be close to the mid-coronal plane, while maximising visibility of seven representative subcortical structures: lateral ventricle, thalamus, caudate, putamen, pallidum, hippocampus and amygdala. The second set consists of dense segmentations of 36 brain structures, estimated from the FLAIR scans using SAMSEG~\cite{Puonti:2016aa}. Leaving aside two cases in which SAMSEG failed, we used these segmentations as a silver standard to evaluate the methods using every available voxel. Moreover, the cerebral tissue labels from these segmentations were also used to simulate a 3D surface scan 
(which in the future we plan to achieve with an inexpensive device)
for hard reconstruction. We also tested a version with a soft reference, using the LPBA40 atlas~\cite{Shattuck:2008aa}.

%{\color{red}
%note: now have final demographic information - last is female with underlying mental health issues - 22 yo may need to re-do median age.}
% Demographic information is provided for this dataset, showing there to be 8 female and 16 male brains with an age range of 22 to 96 years and a median age of 68.
% Cognitive status and clinical diagnosis was also provided for 16 cases giving a split into 4 female dementia cases, 4 male dementia cases and 8 male controls.
% Diagnoses accompanying a decreased cognitive status include frontotemporal dementia, ALS, Alzheimers disease, Parkinson's disease dementia and dementia with Lewy bodies.

\subsection{Experimental Setup}
\label{sec:experiments}

Two volumes were reconstructed, using  hard and soft references respectively, for each case in the dataset.
Reconstructions were computed at 0.5~\si{mm} in-plane resolution, and segmented using the  Bayesian  method  in Section~\ref{sec:segmentation}.
%Reconstructions were computed at $0.5\times0.5\times4$~\si{mm} resolution (coronal), and segmented using the  Bayesian  method  in Section~\ref{sec:segmentation}.
Since direct evaluation of registration error is very difficult, we used two measures of segmentation quality as surrogates. 
First, Dice scores were calculated against manual delineation of cerebral cortex,  white matter, and  the seven  subcortical structures listed in Section~\ref{sec:data}. 
The evaluation was then extended to the whole brain by computing the correlation (and associated $p$ value) between the volumes of the nine structures, derived from the \emph{ex vivo} MRI and from the 3D reconstructed photographic volumes with our method.

\subsection{Results}
\label{sec:resultsSubsec}

\begin{figure}[t]
\centering
\includegraphics[width=\textwidth]{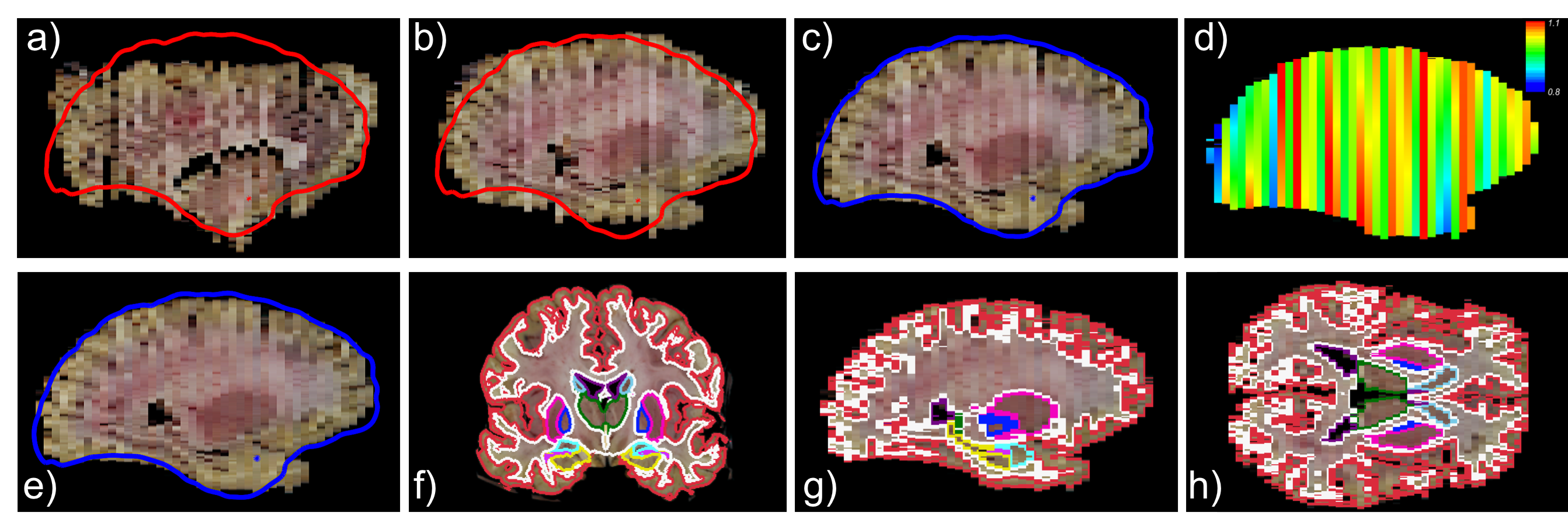}
\caption{
Demonstration of 3D reconstruction and segmentation with a  hard reference. 
The stack of slices~(a) is aligned by COG (b) and registered~(c).
A brightness correction~(d) is calculated and applied to generate a  corrected volume~(e).
Orthogonal views are shown of resulting segmentation labels (f-h). The surface of the reference is represented with red and blue contours  (initialised and registered, respectively). 
}\label{fig:segmentation}
\end{figure} 

\begin{table}[b]
\caption{Correlation coefficient ($r$) and associated $p$ values for the volumes derived from the \emph{ex vivo} MRI with SAMSEG and from the photographs with our method.}\label{tab:correlations}
\begin{tabular}{|c|c|c|c|c|c|c|c|c|c|}
 \hline
Structure&Wh.Ma.&Cortex&Lat.Vent.&Thal.&Caud.&Put.&Pallid.&Hippo.&Amyg\\
\hline
$r$ (hard)&0.80&0.92&0.73&0.78&0.79&0.77&0.63&0.45&0.82\\
$p$ (hard)&$< 10^{-4}$&$< 10^{-8}$&$< 10^{-3}$&$< 10^{-4}$&$< 10^{-4}$&$< 10^{-4}$&$< 0.005$&$< 0.05$&$< 10^{-5}$\\
\hline
$r$ (soft)&0.80&0.84&0.73&0.77&0.81&0.77&0.70&0.28&0.71\\
$p$ (soft)&$< 10^{-4}$&$< 10^{-4}$&$< 10^{-3}$&$< 10^{-4}$&$< 10^{-5}$&$< 10^{-4}$&$< 10^{-3}$&$< 0.2$&$< 10^{-3}$\\
\hline
\end{tabular}
\end{table}

\begin{figure}[t]
\centering
\includegraphics[width=0.27\textwidth]{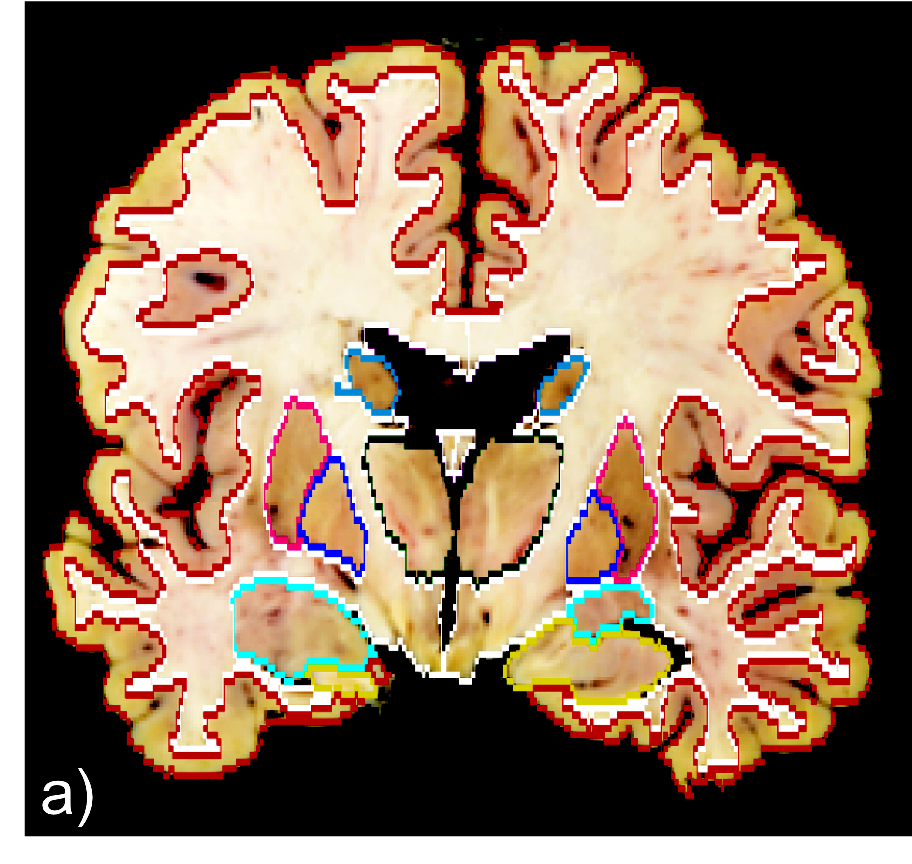}
\includegraphics[width=0.27\textwidth]{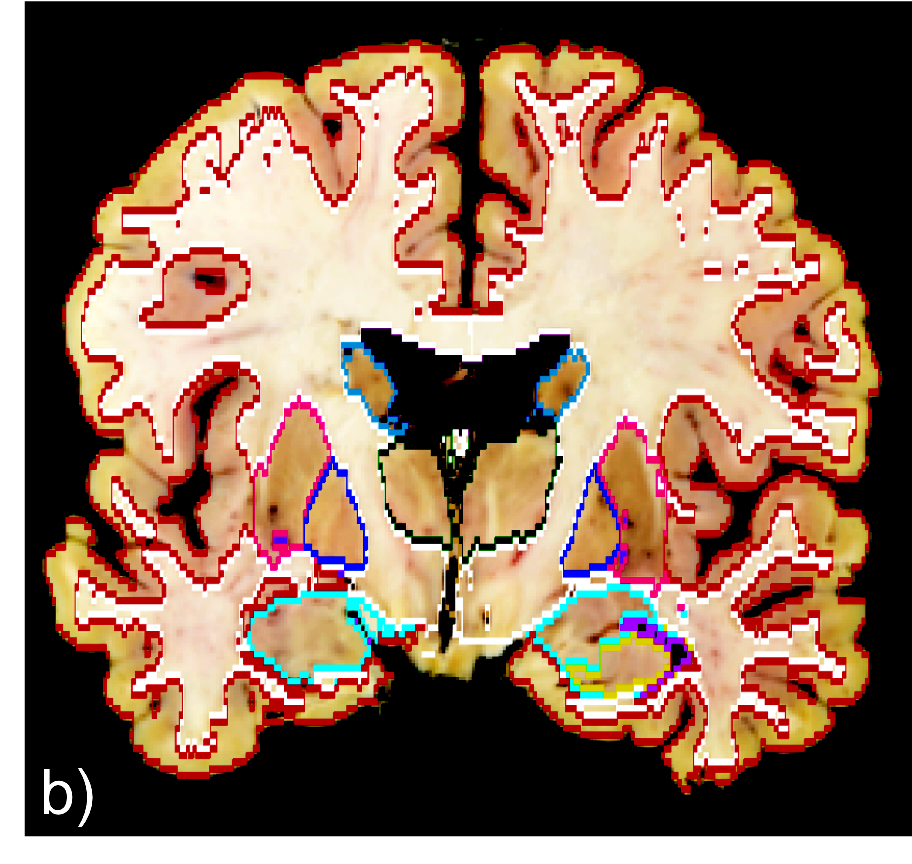}
\includegraphics[width=0.27\textwidth]{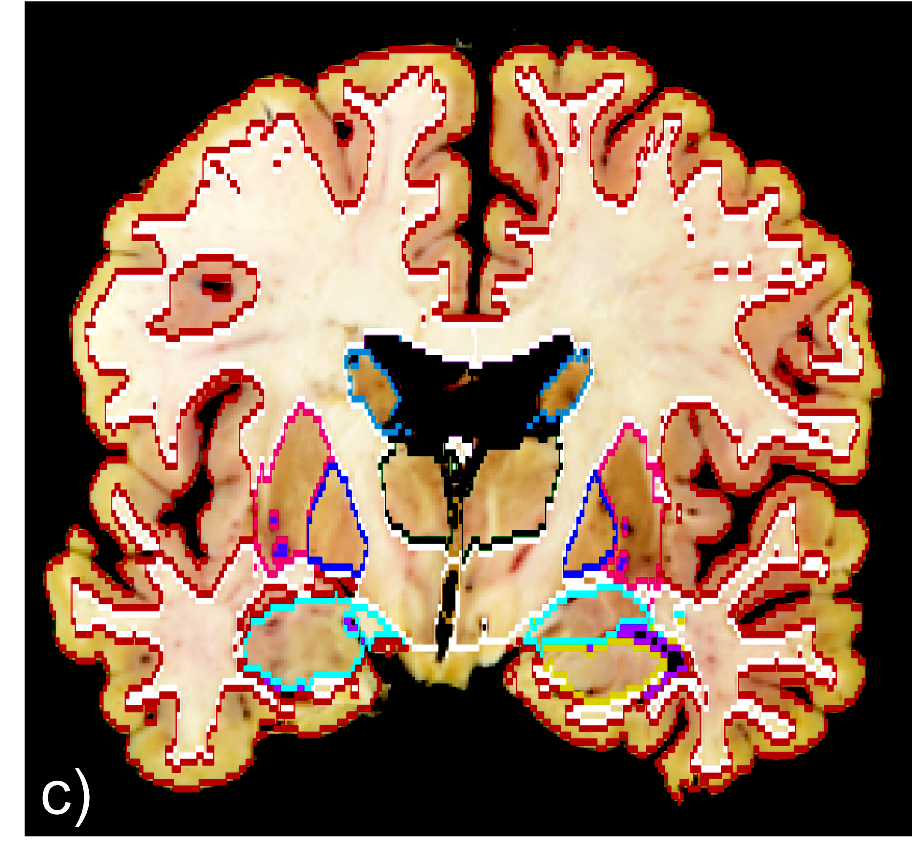}\\\
\includegraphics[width=0.41\textwidth]{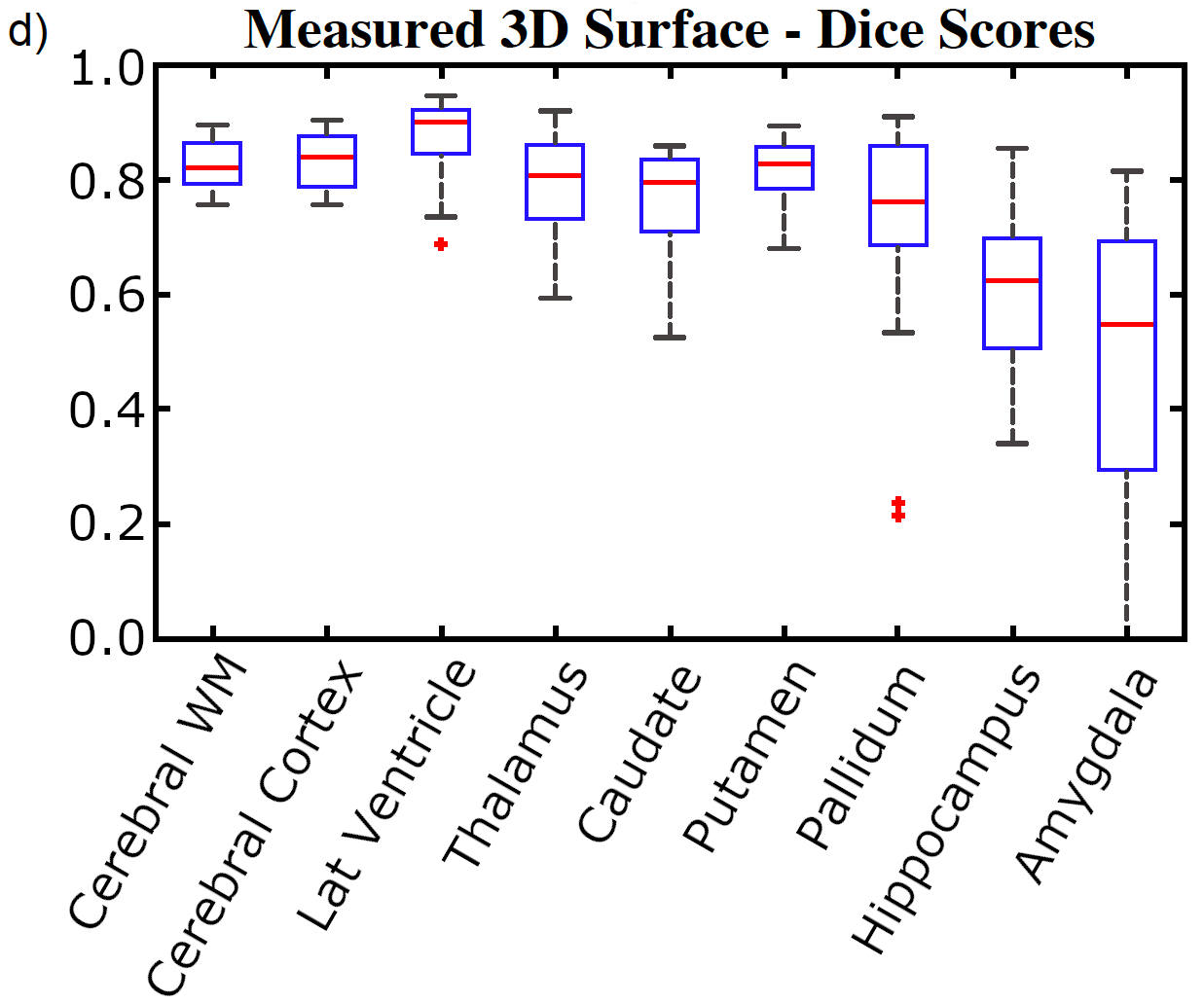}
\includegraphics[width=0.41\textwidth]{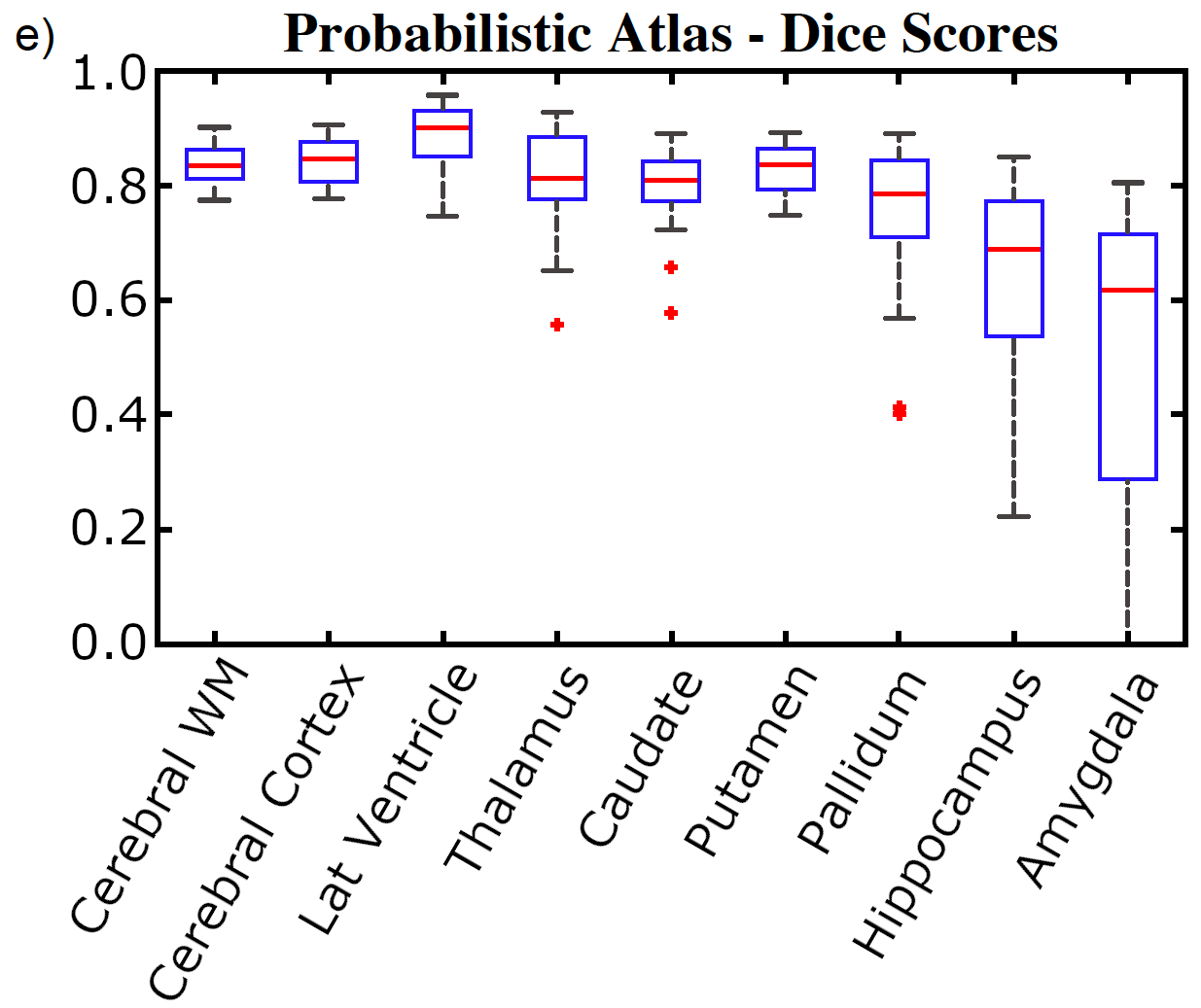}
\caption{
Comparison of manual labelling with proposed automated method. (a)~Manual tracing. (b)~Bayesian segmentation with hard reference. (c)~With soft reference. (d)~Box plots for Dice scores of hard reference using 24 coronal slices. (e)~For soft reference.
}
\label{fig:dice}
\end{figure}

Figure~\ref{fig:segmentation} shows representative images from each stage in the registration and segmentation process, using a hard reference. The proposed procedure successfully aligns the photographs to the reference surface and estimates a brightness field that clearly increases the homogeneity of the image intensities, enabling accurate segmentation. Further qualitative results are shown in  Figures~\ref{fig:dice}(a-c), which compare the manual delineations with our automated segmentations, using both the hard and  soft references. The  corresponding quantitative results (Dice scores) are shown in Figures~\ref{fig:dice}(d-e). The segmentations are quite accurate for most structures, except for the hippocampus and amygdala, whose interface is difficult to separate in this particular coronal plane. The method also commits minor mistakes that are common in Bayesian segmentation, e.g., including the claustrum  in the putamen. But overall, the Dice scores are competitive (above 0.8 for many structures), which is very encouraging given that they are computed from photographs. 
Particularly high scores are achieved for the cerebral cortex, since there is no extracerebral tissue in the images.

While Dice scores on a single slice are informative, in order to show that  measured trends from photo volumes are transferable to clinical imaging modalities it is crucial to test whether the volumes computed with our method on thick slices correlate well with the volumes derived from the isotropic MRI. 
For this reason we compare the silver standard volumes derived from the \emph{ex vivo} MRI, with the volumes given by our proposed method. 
Table~\ref{tab:correlations} shows the correlation coefficients and associated $p$ values for the nine representative structures of interest. The results are consistent with the Dice scores on the sparse slices, showing strong correlations and significance for all structures except for the pallidum and the hippocampus. The pallidum is notoriously difficult to segment, even in MRI, due its low contrast. The hippocampus seems to be particularly affected by the large slice thickness in our reconstructed photography volumes.

\section{Discussion and conclusion}
\label{sec:discussion}

We have presented the first algorithm for the construction of registered dissection photography volumes using only an external boundary shape as reference. For this proof of concept study we tested our methods on 24 cases, and assessed accuracy and sensitivity using Dice scores and correlations to silver standard volumes, respectively. The results are promising, both with the hard and the soft reference, and pave the way for inexpensive, large-scale NTNC studies -- even retrospectively, using the soft version. 

Future work will follow several directions. 
First, we will  more thoroughly validate the methods, using additional cases, metrics, and manually traced images. 
We also plan to extend our method to slices with uneven thickness, and explore imputation algorithms (e.g.,~\cite{Dalca:2019aa}) to increase the resolution of the 3D reconstructed scans. Access to super-resolved isotropic volumes is expected to enhance the quality of the segmentations (e.g., for the hippocampus), and also has the potential to enable other volumetric analyses that underperform with insufficient resolution (e.g., registration,  cortical thickness). These additional analyses will likely benefit from a hard external reference: while drifts in the 3D reconstruction towards an average shape due to the probabilistic atlas do not seem to  penalise segmentation, we hypothesise that using a 3D surface scan (an increasingly inexpensive technology) like the one in Fig.~\ref{fig:workflow} will increase the precision of cortical measurements and registration, and enable the discovery of new imaging markers to study neurodegenerative diseases. 
 
\subsubsection{Acknowledgements}
Work primarily supported by ARUK (ARUK-IRG2019A-003) and the ERC (677697).
Additional support provided in part by:  
the BRAIN Initiative (U01-MH117023);
NIH (P50-AG005136,  U01-AG006781, 5U01-MH093765);
NIMH;
NIBIB (P41-EB015896, 1R01-EB023281, R01-EB006758, R21-EB018907, R01-EB019956);  
NIA (1R56-AG064027, 1R01-AG064027, 5R01-AG008122, R01-AG016495);
NIDDK (1-R21-DK-108277-01);
NINDS (R01-NS0525851, R21-NS072652, R01-NS070963, R01-NS083534, 5U01-NS086625, 5U24-NS10059103, R01-NS105820, R01-NS112161);
HU00011920008 subaward to CMD;  
the Nancy and Buster Alvord Endowment;
the EU H2020 (Marie Sklodowska-Curie grant 765148);  
 Shared Instrumentation Grants 1S10RR023401, 1S10RR019307, 1S10RR023043. 
BF has a financial interest in CorticoMetrics, %, a company whose medical pursuits focus on brain imaging and measurement technologies. 
managed by MGH and Partners HealthCare.
Authors would also like to thank L.~Keene, K.~Kern, A.~Keen and E.~Melief for their assistance with data acquisition.

%\jei{if accepted, in the final version, acknowledgements would go here  (ERC, ARUK, etc)}

\bibliographystyle{splncs04}
\bibliography{paper307}

\end{document}